\documentclass{article} 
\usepackage{iclr2026_conference,times}

\usepackage{amsmath,amsfonts,bm}

\def\eqref#1{equation~\ref{#1}}

\def\1{\bm{1}}

\DeclareMathAlphabet{\mathsfit}{\encodingdefault}{\sfdefault}{m}{sl}
\SetMathAlphabet{\mathsfit}{bold}{\encodingdefault}{\sfdefault}{bx}{n}

\usepackage{url}
\usepackage{xspace}
\usepackage{graphicx}
\usepackage{tabularx}
\usepackage{multirow}
\usepackage{wrapfig}
\usepackage{comment}
\usepackage{booktabs}       
\usepackage{caption}
\definecolor{citecolor}{HTML}{0071bc}
\usepackage[table]{xcolor} 
\definecolor{shadecolor}{gray}{0.9} 
\usepackage[colorlinks=true, linkcolor=red, anchorcolor=blue, 
            citecolor=citecolor, pagebackref=true]{hyperref}
\usepackage{cleveref}
\title{PhysMaster: Mastering  Physical Representation for Video Generation via Reinforcement Learning}

\author{Sihui Ji$^{1^{\ast}}$, 
Xi Chen$^{1}$, 
Xin Tao$^{2}$,
Pengfei Wan$^{2}$,
Hengshuang Zhao$^{1^{\dagger}}$
\\
$^{1}$The University of Hong Kong\quad
$^{2}$Kling Team, Kuaishou Technology
}

\iclrfinalcopy 
\newcommand\nnfootnote[1]{%
  \begin{NoHyper}
  \renewcommand\thefootnote{}\footnote{#1}%
  \addtocounter{footnote}{-1}%
  \end{NoHyper}
}

\newcommand{\method}{PhysMaster\xspace}
\newcommand{\encoder}{PhysEncoder\xspace}

\begin{document}

\maketitle

\nnfootnote{$\ast$ Work done during an internship at Kling Team, Kuaishou Technology. $\dagger$ Corresponding author.}

\begin{center}
    \vspace{-47pt}
    \textbf{\url{https://sihuiji.github.io/PhysMaster-Page/}}
    \vspace{5pt}
\end{center}%

\begin{abstract}

Video generation models nowadays are capable of generating visually realistic videos, but often fail to adhere to physical laws, limiting their ability to generate physically plausible videos and serve as ``world models''.
To address this issue, we propose \method, which captures physical knowledge as a representation for guiding video generation models to enhance their physics-awareness.
Specifically, \method is based on the image-to-video task where the model is expected to 
predict physically plausible dynamics from the input image. 
Since the input image provides physical priors like positions, materials, and interactions of objects in the scenario, we devise \encoder to encode such physical representation as an extra condition. 
Nevertheless, there is no well-established definition for a physical representation. Thus, we could not use an off-the-shelf model to extract the physical condition, or set a straightforward supervision to train \encoder. 
To solve this challenge, we adopt a top-down optimization strategy, where the \encoder is optimized based on the physical plausibility of the final generated videos using reinforcement learning~(RL). 
Through this top-down optimization, the \encoder can effectively capture implicit physical cues from the starting image and inject them into the video generation process.
Experiment results prove that, our method significantly enhances the model’s physics-awareness as a plug-in, and demonstrate strong performance on both specialized proxy tasks and general open-world scenarios.

\end{abstract}

\section{Introduction}
\label{sec:intro}
Video generation models~\citep{videoworldsimulators2024, kling, yang2024cogvideox} have developed rapidly nowadays, achieving significant performances in generating 
visually appealing videos~\citep{gen3,genmo2024mochi,kong2024hunyuanvideo}.
However, they primarily act as sophisticated pixel predictors based on case-specific imitation, and often face challenges in adherence to physical laws~\citep{kang2024far, liu2025generative, meng2025grounding}.
This limits their ability to generate physically plausible videos and 
further comprehend physical principles to serve as ``world models''.
To evolve these models from content creators to world simulators, we aim to 
incorporate physical knowledge into the video generation process to enhance their physical realism.

We summarize the specific challenges of physics-aware video generation into the following two points. First, the commonly used Mean Squared Error~(MSE) loss for data-driven finetuning focuses on appearance fitting rather than comprehension of physical knowledge, and it is non-trivial to directly supervise the physical performance of pretrained models beyond merely appearance.
Second, generative models struggle to extract appropriate physical knowledge from a textual instruction or an input image and translate it into physical guidance for generation, which demands logical reasoning from descriptions or images to physical knowledge, and to visual phenomena. Progress for enhancing physics-awareness of video generation has been made and the solutions can be broadly categorized into two types based on the usage of simulation. Simulation-based approaches~\citep{lv2024gpt4motion, liu2024physgen} attempt to apply physics-based simulation results to guide video generation, but they are often 
constrained in the range of simulable physical processes and modalities, lacking the potential to generalize to diverse phenomena. 
Simulation-free methods~\citep{xue2024phyt2v, furuta2024improving} rely on post-training on physics-rich data or employ reinforcement learning for aligning to human preference. The former highly depends on fitting similar training samples, and the latter utilizes either expensive human annotation suffering from rater variability, or scalable but inaccurate AI evaluators.
In summary, existing works find it hard to truly abstract and understand physics of the world~\citep{lin2025exploring, motamed2025generative}, hindering generalization to diverse physics.


Facing the aforementioned challenges, we propose to learn a physical representation as the bridge between the necessary physical knowledge and generated videos to guide generative models towards physics awareness.
Specifically, we focus on image-to-video~(I2V) generation where an initial frame and textual description are given, the model is supposed to predict physically plausible dynamics from input scenes. 
The input image offers visual cues like object configurations, relative positions, and potential interactions that largely dictate the subsequent physical evolution of video, making it 
a reliable source of physical priors. Thus firstly, we devise a physical encoder, \encoder, to extract implicit physical representation from the input image as an extra input condition to guide the generation process for enhancing physics-awareness of the model.

However, how to learn an effective physical representation for video generation remains an open question. Without a well-established definition for physical representation, we can not conduct straightforward supervision of \encoder for training. Thus we propose a top-down optimization strategy to optimize its ability for guiding physically plausible video generation by reinforcement learning with human feedback~(RLHF) framework, which has proven its effectiveness in finetuning of both large language models~(LLMs)~\citep{yuan2023rrhf, xu2024contrastive, yuan2023rrhf} and generative models~\citep{lee2023aligning, prabhudesai2024video, fan2023dpok}.
The physical plausibility of generated videos guided by \encoder acts as feedback for optimizing \encoder to extract effective physical representations.
Specifically, we train \encoder on human preference data via Direct Preference Optimization~(DPO)~\citep{rafailov2023direct} in a three-stage training pipeline. We first conduct supervised fine-tuning~(SFT) of both base model and \encoder, then we adopt a two-stage DPO
with pairwise supervision based on the physical plausibility of the final generated
videos to enhance \encoder's capacity to capture physical representations and model's physical performance, with trainable module separately set as 
LoRA~\citep{hu2021lora} of the DiT model and \encoder. With SFT providing the model with initial ability to predict 
physically plausible videos under the guidance of simultaneously finetuned \encoder, the subsequent DPO processes further enable \encoder to effectively capture implicit physical cues
from the starting image and inject them into video generation, thus helping improve physical understanding of the model.

Last but not least, we condition the video generation model on physical representation in a plug-in manner, enhancing physics-awareness by injecting physical knowledge into it. Such a paradigm enables the model to learn general physical properties, rather than overfitting to specific phenomena or being constrained to particular motion modalities as in previous works, thus allowing it to generalize to diverse scenarios.
We demonstrate the effectiveness of \encoder starting from a specialized proxy task of ``free-fall'', and then generalize to general open-world physical scenarios governed by a wide range of physical laws. \encoder proves its capablity by guiding the generation model towards enhanced physical performance, and such generalization implies that \method, our representation learning paradigm facilitates the physical comprehension in a broad scope.

In summary, \method provides a more generalizable solution for video generation models to capture physical knowledge across diverse physical phenomena, 
showing its advantage in acting as a foundational solution for physics-aware video generation and potential to energize more fancy applications~\citep{agarwal2025cosmos, yang2024video, yang2023learning}.

\section{Related Works}
\label{sec:related_works}
\noindent\textbf{Physics-aware video generation.}
While recent video generation models achieve impressive visual effects~\citep{videoworldsimulators2024, kong2024hunyuanvideo}, they still struggle with adherence to real-world physical laws~\citep{lin2025exploring, kang2024far}. 
Physics-aware video generation approaches can be broadly categorized based on the application of explicit physical simulation. Simulation-based methods~\citep{lv2024gpt4motion, xie2025physanimator, montanaro2024motioncraft, zhang2024physdreamer} guide generation with 
simulation results.
PhysGen~\citep{liu2024physgen} utilizes rigid-body dynamics simulated with physical parameters inferred by large foundation models. PhysMotion~\citep{tan2024physmotion} relies on MPM-based simulation to generate coarse videos which are refined by a video diffusion model. 
As for simulation-free approaches, they either fine-tune on large-scale video datasets to implicitly internalize physical priors~\citep{wang2025wisa, zhang2025think}, or use reinforcement learning with feedback from human annotators or vision-language models~\citep{xue2024phyt2v,furuta2024improving}. 
PhyT2V~\citep{xue2024phyt2v} uses MLLMs to refine prompts iteratively through multiple rounds of generation and reasoning.
WISA~\citep{wang2025wisa} incorporates structured physical information into the generative model and uses Mixture-of-Experts for different physics categories.
However, those methods are restricted to fixed physical categories or exhibit limited physical comprehension. Our \method incorporates physical knowledge into video generation process via physical representation to enhance general physics-awareness.

\noindent\textbf{RLHF for video generation.}
Inspired by the success of RLHF in LLMs~\citep{ouyang2022training, jaech2024openai}, researchers have explored applying this paradigm to video generation~\citep{zhang2024onlinevpo, qian2025rdpo}. VideoDPO~\citep{liu2024videodpo} pioneers the adaptation of DPO~\citep{rafailov2023direct} to video diffusion models by considering both visual quality and semantic alignment for data pair construction. VideoAlign~\citep{liu2025improving} introduces a multi-dimensional video reward model and DPO for flow-based video generation model based on it.
PISA~\citep{li2025pisa} investigates specifically for video generation of object free-fall, improving physical accuracy through reward modeling based on depth and optical flow. 
Unlike the aforementioned methods, we optimize a physical encoder rather than the whole video generation model by leveraging generative feedback from the model's outputs.
This paradigm mitigates overfitting to specific physical processes and promotes the encoder’s generalizability for learning universal physical knowledge through RLHF.

\section{Method}
\label{sec:method}

Based on I2V setting, \method extracts physical representation from the input image and optimizes both the generation model and \encoder in a three-stage training pipeline.
It seeks direct supervision from groundtruth via SFT and pairwise supervision from generated videos via DPO, and is implemented on the specialized proxy task and general open-world scenarios.
We will separately detail physical representation~(Sec~\ref{sec:representation}), task formulation~(Sec~\ref{sec:task_formulation}) and training scheme~(Sec~\ref{sec:training}).

\begin{figure*}[t]
\centering 
\includegraphics[width=1.0\linewidth]{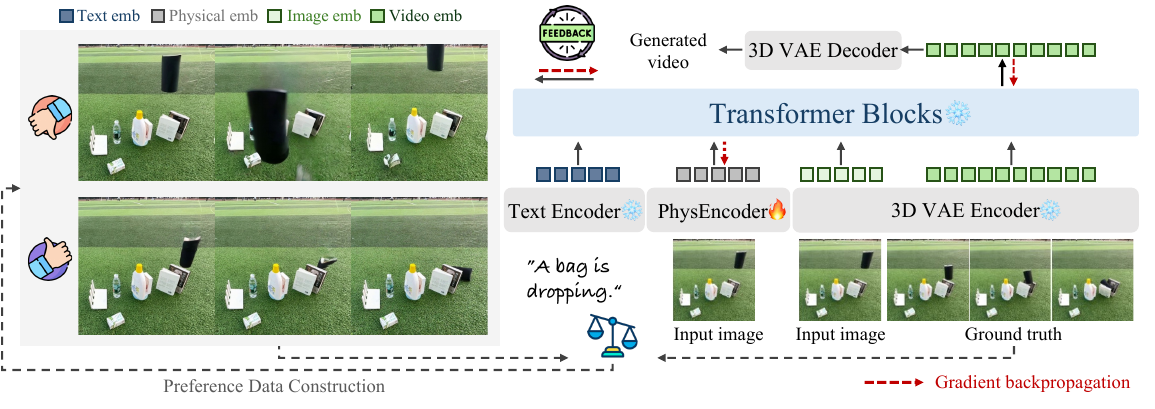} 
\vspace{-18pt}
\caption{%
    \textbf{Overall architecture of \method.}
    Given an input image, \encoder encodes its
physical feature and concatenates with visual features, then the DiT model predicts subsequent frames conditioned on physical, visual, and text embeddings. We optimize \encoder’s physical representation via feedback from generated video pairs of the model
by maximizing reward derived from ``positive'' and ``negative'' video outputs in a DPO paradigm.
}
\vspace{-15pt}
\label{fig:pipeline}
\end{figure*}

\subsection{Physical Representation}

\label{sec:representation}
\method is implemented upon a transformer-based diffusion model~(DiT)~\citep{peebles2023scalable}, which employs 3D Variational Autoencoder~(VAE)~\citep{kingma2013auto} 
to transform videos and initial frame to latent space, and T5 encoder $\mathcal{E}_{T5}$~\citep{raffel2020exploring} for text embeddings $c_{text}$.

We propose to learn a physical representation from input image as extra guidance for the I2V model to inject physical information, since the input image contains not only explicit physical states, such as object material and spatial distribution, but also implicit physical laws, like the gravitational field.
It is worth expecting that the learned physical representation can be used as a generalizable guidance of both physical properties and dynamics for physics-aware video generation.
Following the structure of Depth Anything~\citep{yang2024depth}, we build \encoder with a DINOv2~\citep{oquab2023dinov2} encoder and a physical head. The former adopts pretrained weights from~\citet{yang2024depth} for initialization and
takes the role of semantic perception, while the latter adapts the extracted high-level semantic features into an appropriate dimension to be injected into the DiT model. 
Taking the first frame as image input, \encoder encodes it into physical embeddings $c_{phys}$, which are then fed into DiT model after concatenated with image embeddings $c_{image}$.
For SFT, the flow-based DiT model with weights ${\theta}$ directly parameterizes the $v_{\theta}(z_t, t, c_{text},c_{image}, c_{phys})$ to regress velocity $(z_1 - z_0)$ with the Flow Matching objective~\citep{lipman2022flow}:
\begin{equation}\label{eq:loss_sft}
    \mathcal{L}_{LCM}=\mathbb{E}_{t,z_0,\epsilon} ||v_{\theta}(z_t,t,c_{text},c_{image}, c_{phys})-(z_1-z_0)||_2^2.
\end{equation}

\subsection{Task Formulation}
\label{sec:task_formulation}

Our work aims to provide a scalable and generalizable methodology for learning physics from targeted data, so for demonstrating the effectiveness of our \method, we start by defining a proxy task under specialized physical principles and construct domain-specific data for preliminary validation; then we verify its generalizability across a broader range of physical laws and various tasks.

\textbf{Proxy task.} For preliminary verification, ``free-fall''~(involving the complete physical process of objects dropping from mid-air and colliding with other objects on a surface), a specialized yet expressive scenario is chosen as the proxy task for the following characteristics. First, ``free-fall'' embodies clear and fundamental physical principles~(e.g., energy and momentum conservation) shared across diverse physical scenarios, making it a suitable representative for further generalization. Second, such physical scenario involves a wide range of object-level physical properties, such as density, elasticity, and hardness, allowing proof of generalizability of learned representations across different physical attributes. Third, this task can be easily simulated for scalable generation of synthetic data and allows for straightforward evaluation by comparing generated videos against ground-truths. The reason is that by assuming the falling object starts from rest and is only influenced by gravity, the trajectories of objects become fully deterministic given the initial frame, which also enables automatic construction of preference video pairs for DPO by similarity with ground-truths.





\textbf{General open-world scenarios.} We further substantiate generalization capabilities of \method across diverse physical processes. Following WISA~\citep{wang2025wisa}, we include large-scale scenarios broadly covering common physical phenomena observed in real world for \encoder to acquire a far more comprehensive and generalizable understanding of physical laws and thus effectively enhances the physics awareness of the video generation model. Different from the proxy task implementation, we modify the text prompts provided to the generation model by adding domain-specific prefixes~(e.g., ``\textit{Optic}, A ray of light ...'', ``\textit{Thermodynamic}, A glass of water ...''). This conditions the model on the type of involved physics laws and guides it to associate visual phenomena with underlying physics extracted from PhysEncoder. For preference assignment, we rely on human annotators to provide pairwise labels for DPO data construction and evaluation.

\subsection{Training Scheme}
\label{sec:training}
We propose a three-stage training pipeline for \method to enable physical representation learning of \encoder by leveraging the generative feedback from I2V model. The core idea is formulating DPO for \encoder with the reward signal from generated videos of pretrained DiT model, 
thus help physical knowledge learning without explicit modeling.

\textbf{Stage I: SFT for DiT and \encoder.} First, we condition the I2V base model on physical representation from \encoder by SFT, thus it is possible for us
to optimize \encoder with the performance of model as feedback in following stages. Since \encoder's training starts from the frozen DINOv2 with pretrained weights from Depth Anything~\citep{yang2024depth} and trainable physical head with randomly initialized weights, this stage can be viewed as adapting Depth Anything for physical condition injection.
As in Figure~\ref{fig:pipeline}, by concatenating physical embeddings extracted by \encoder with visual embeddings encoded by VAE,
we inject physical representation as extra condition to the model. 
SFT following Eq~\ref{eq:loss_sft} equips the model with the initial
ability to predict subsequent frames from the input image, guided by simultaneously finetuned \encoder.

\textbf{Stage II: DPO for DiT.}
Second, we expect to adapt the output of the pretrained model to a more physically plausible distribution, paving the
way for the \encoder to learn from generated videos with higher physical accuracy.
Then in Stage II, we apply LoRA~\citep{hu2021lora} to finetune the DiT model on preference dataset with DPO, during which the model learns to generate positive samples with higher probability and negative samples with lower probability.
Regarding I2V setting, each sample in our preference dataset includes a prompt $p$, an image $i$, a human-chosen video 
$x^w$ and a human-rejected video $x^l$.
The goal of DPO is to learn a conditional distribution \(\pi_\theta(x \mid p, i)\) that maximizes 
the reward $r_\phi (x,p,i)$ while staying close to a reference model \(\pi_{\text{ref}}\):

\begin{align}\label{eq:rlhf}
    \max_{\pi_\theta} \ & \mathbb{E}_{p \sim \mathcal{D}_p, i \sim \mathcal{D}_i, x \sim \pi_\theta(x \mid p,i)} \left[ r_\phi (x,p,i) \right]
    - \beta \, \mathbb{D}_{\text{KL}} \left[ \pi_\theta(x \mid p,i) \,\|\, \pi_{\text{ref}}(x \mid p,i) \right]
\end{align}

where 
$\beta$ controls the regularization term~(KL-divergence) from \(\pi_{\text{ref}}\).
For our flow-based DiT model, Flow-DPO objective~\citep{liu2025improving} \(L_\text{FD}(\theta)\) is then given by:

\begin{small}
\begin{align}\label{eq:flow_dpo}
- \mathbb{E} \Bigg[ \log \sigma \Bigg(& -\frac{\beta}{2} \Big( 
    \| v^w - v_\theta(x_{t}^w, t)\|^2 
    - \|v^w - v_\text{ref}(x_{t}^w, t)\|^2
    - \bigl(\|v^l - v_\theta(x_{t}^l, t)\|^2 
    - \|v^l - v_\text{ref}(x_{t}^l, t)\|^2 \bigr) \Big) \Bigg) \Bigg],
\end{align} 
\end{small}
where the conditioning prompt $p$ and image $i$ are omitted for simplicity, \(v_\theta\) denotes the predicted velocity field,  \(v^w\), \(v^l\) are the target velocity of ``preferred'' and ``less preferred'' data.

The pretrained DiT model from Stage I is regarded as the reference model and is used to construct the preference data pairs. Specifically, we generate two groups of videos with the pretrained model using the same prompt $p$ and initial frame $i$ but different seeds. 
By establishing clear distinctions between positive samples $x^w$ and negative samples $x^l$, the model learns to generate physically plausible
videos. As a result, we further enhance the model's physics awareness.

\textbf{Stage III: DPO for \encoder.} We leverage generative feedback from the pretrained DiT model to optimize \encoder's physical representation via DPO paradigm.
As illustrated in Figure~\ref{fig:pipeline}, our framework consists of two parts: \encoder to be optimized and the pretrained DiT model providing generative feedback. 
With physical head of \encoder the only trainable module, Stage III shares the same training objective Eq~\ref{eq:flow_dpo} with Stage II, differing solely in the learnable parameters.
\(L_\text{FD}(\theta)\) leads \encoder to learn a physical representation that
guides the predicted velocity field $v_\theta$ closer to the target velocity $v^w$ of the ``preferred'' data.
In this manner, by directing the DiT 
model to generate more accurate physical dynamics, the \encoder's original representation is gradually 
optimized with more physical knowledge through
model feedback. 

\section{Experiments}
\label{sec:exp}

To evaluate the effectiveness of \method for physical representation learning 
and demonstrate its potential to enhance physical performance of the DiT model, comprehensive experiments are conducted 
on both the proxy task and wide-ranging scenarios.

\subsection{Implentation Details}
\label{sec:implementation_details}
\noindent \textbf{Training configuration.} The training of both \encoder and the DiT model is conducted on 8 NVIDIA-A800 GPUs in all three stages, with 20 hours for SFT, 15 hours for DPO on LoRA and 8 hours for DPO on \encoder. 
The training process employs the Adam optimizer~\citep{kingma2014adam}, and we utilize 50 DDIM steps~\citep{song2020denoising} 
and set the CFG scale to 7.5 during inference.

\noindent \textbf{Dataset construction.} 
For the proxy task, we follow PISA~\citep{li2025pisa} to use Kubric~\citep{greff2022kubric} to create synthetic datasets of ``free-fall''. The object assets are sourced from the Google Scanned Objects~(GSO) dataset~\citep{downs2022google}. For generalizability demonstration, we utilize WISA-80K~\citep{wang2025wisa} encompassing 17 types of real-world physical events across three major branches of physics~(Dynamics, Thermodynamics, and Optics).

\noindent \textbf{Evaluation protocols.}
PisaBench~\citep{li2025pisa} is introduced to evaluate our model's performance on the proxy task. 
We use SAM 2~\citep{ravi2025sam} for segmentation of object masks and compute the following metrics between corresponding masks of generated and ground truth videos for evaluation: L2 distance between the centroids of the masked regions, chamfer distance~(CD) and Intersection over Union~(IoU) of the mask regions.
We utilize VIDEOPHY~\citep{bansal2024videophy} for evaluating physics awareness of video generation in general open-world scenarios. We test on 344 carefully crafted prompts from it, which reflect a wide array of physical principles, and report the physical commonsense~(PC) and semantic adherence~(SA) scores.

\subsection{Evaluation on Proxy Task}
\label{sec:evaluation}

\begin{table*}[t]
\setlength{\belowcaptionskip}{0cm}
\caption{\textbf{Ablation study for models from different training stages and strategies} on proxy task, evaluated on the test set split into ``seen'' and ``unseen''. $v_\theta$ is DiT model, $E_{p}$ is \encoder.
}
\label{tab:abla_freefall}
\vspace{-5pt}
\centering

\resizebox{1\textwidth}{!}{ 
    \begin{tabular}{lccc ccc ccc}
    \toprule
    \multirow{2}{*}{Training Stages} & \multicolumn{3}{c}{Seen} & \multicolumn{3}{c}{Unseen} & \multicolumn{3}{c}{Average} \\
    \cmidrule(lr){2-4} \cmidrule(lr){5-7} \cmidrule(lr){8-10} 
    & L2~($\downarrow$) & CD~($\downarrow$) & IoU~($\uparrow$) & L2~($\downarrow$) & CD~($\downarrow$) & IoU~($\uparrow$) & L2~($\downarrow$) & CD~($\downarrow$) & IoU~($\uparrow$) \\
    \midrule
    Base & 0.1066 & 0.323 & 0.119 & 0.1065 & 0.339 & 0.111 & 0.1066 & 0.331 & 0.115 \\
    \midrule
    SFT for $v_\theta$ & 0.0532 & 0.134 & 0.137 & 0.0512 & 0.133 & 0.135 & 0.0522 & 0.134 & 0.136 \\
    SFT for $v_\theta$ \& $E_p$~(Stage I) & 0.0568 & 0.141 & 0.137 & 0.0498 & 0.128 & 0.134 & 0.0533 & 0.134 & 0.135 \\
    \midrule
    SFT for $v_\theta$ + DPO for $v_\theta$  & 0.0560 & 0.143 & \textbf{0.144} & \textbf{0.0446} & \textbf{0.115} & 0.128 & 0.0503 & 0.129 & 0.136 \\
    SFT for $v_\theta$ \& $E_p$ + DPO for $v_\theta$~(Stage II) & 0.0520 & 0.125 & 0.133 & 0.0454 & 0.116 & 0.136 & 0.0487 & 0.120 & 0.134 \\
    SFT for $v_\theta$ \& $E_p$ + DPO for $E_{p}$ & 0.0559 & 0.140 & 0.138 & 0.0503 & 0.129 & 0.134 & 0.0531 & 0.134 & 0.136 \\
    SFT for $v_\theta$ \& $E_p$ + DPO for $v_\theta$ \& $E_{p}$ & 0.0501 & 0.124 & \underline{0.141} & 0.0458 & 0.121 & 0.140 & 0.0480 & 0.123 & 0.141 \\
    \midrule
    \rowcolor{shadecolor} 
    SFT for $v_\theta$ \& $E_p$ + DPO for $v_\theta$ + DPO for $E_{p}$~(Stage III) & \underline{0.0489} & \textbf{0.120} & \underline{0.141} & \underline{0.0450} & \textbf{0.115} & \textbf{0.145} & \underline{0.0470} & \textbf{0.118} & \textbf{0.143} \\
    SFT for $v_\theta$ \& $E_p$ + DPO for $v_\theta$ + DPO for $v_\theta$ \& $E_{p}$ & \textbf{0.0477} & \underline{0.121} & \textbf{0.144} & 0.0452 & \textbf{0.115} & \underline{0.141} & \textbf{0.0466} & \textbf{0.118} & \underline{0.142} \\
    \bottomrule
    \end{tabular}
}
\vspace{-9pt}
\end{table*}

\begin{figure*}[t]
\centering 
\includegraphics[width=1\linewidth]{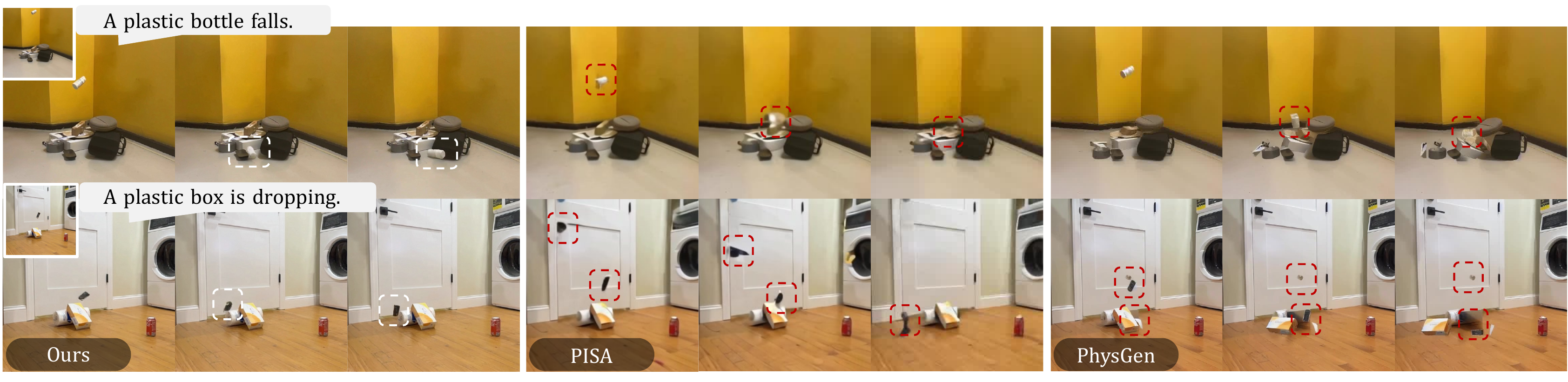} 
\vspace{-15pt}
\caption{%
    \textbf{Qualitative comparison with video generation models specialized for rigid-body motion} proves the advantage of our model in shape consistency and trajectory accuracy on ``free-fall''.}
    
\label{fig:comp_freefall}
\vspace{-15pt}
\end{figure*}

\begin{figure*}[t]
\centering 
\includegraphics[width=1\linewidth]{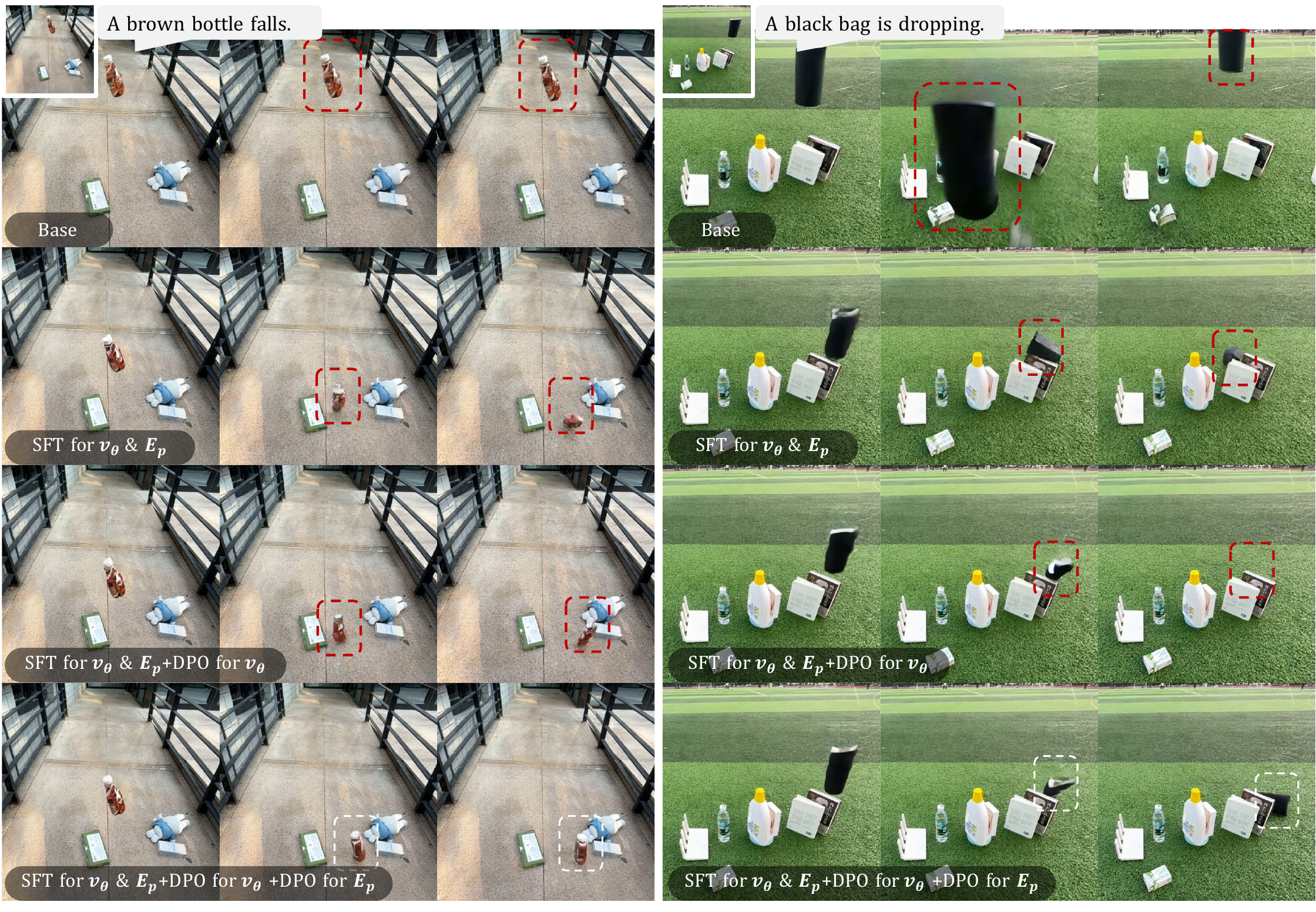} 
\vspace{-15pt}
\caption{%
    \textbf{Qualitative ablation for models in each training stage} on proxy task. The model exhibits a preliminary capability for predicting object motion trends after SFT. Two-stage DPO further improves model performance in preserving objects' rigidity and complying with physical laws~(e.g., gravitational acceleration and collision). $v_\theta$ is DiT model and $E_p$ is \encoder.}
    
\label{fig:stage_real}
\vspace{-6pt}
\end{figure*}

To validate that our training pipeline can effectively improve the physical performance of base model on the proxy task, we compare the physical accuracy of our model on "free-fall" motion with existing works and ablate different training techniques of \encoder.

\noindent \textbf{Comparison.} 
We compare our model with PhysGen~\citep{liu2024physgen} and PISA~\citep{li2025pisa} on the real-world subset from PisaBench~\citep{li2025pisa} which is unseen to any model during training for robust evaluation. We apply more rigorous metrics of similarity over all objects in the scene by comparing against the ground truth.
\begin{figure}[!t]
    \begin{minipage}{0.4\textwidth}
        \centering
        \captionof{table}{\textbf{Quantitative comparison with video generation models specialized for rigid-body motion} verifies superiority of our model on proxy task.}
        \vspace{-5pt}
        \label{tab:comp_freefall}
        \scriptsize
        \setlength{\tabcolsep}{3.3pt}
        \begin{tabularx}{1\textwidth}{@{}lXXX@{}}
            \toprule
            Methods & L2~($\downarrow$) & CD~($\downarrow$) & IoU~($\uparrow$) \\
            \midrule
            PhysGen & 0.0433 & 0.0967 & 0.418 \\
            PISA & \textbf{0.0294} & 0.0570 & 0.433 \\
            \rowcolor{shadecolor} 
            Ours & 0.0299 & \textbf{0.0567} & \textbf{0.468} \\
            \bottomrule
        \end{tabularx}
    \end{minipage}\hfill
    \begin{minipage}{0.58\textwidth}
        \centering
        \includegraphics[width=1\textwidth]{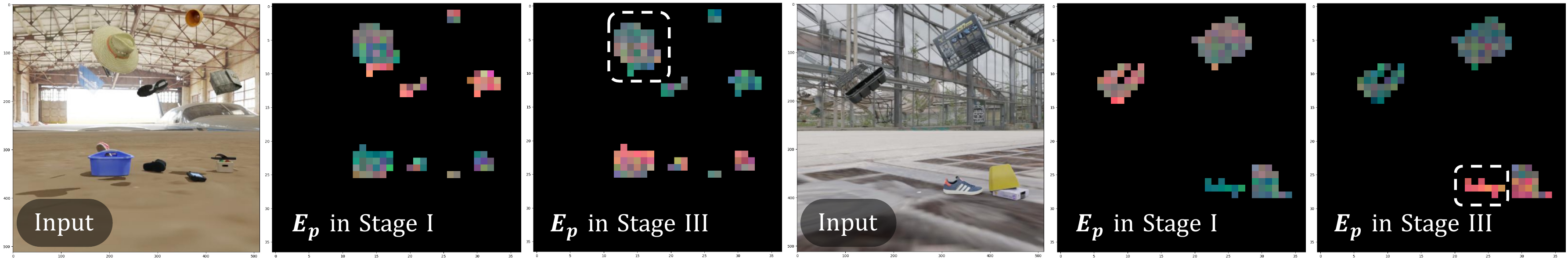} 
        \vspace{-12pt}
        \caption{\textbf{Visualization of the first three PCA components of physical representation.} \textbf{$E_p$ in Stage III} reveals similarities in objects under the same external forces~(red: on the ground; green: in the air) over \textbf{$E_p$ in Stage I}. }
        \label{fig:rep}
    \end{minipage}
    \vspace{-15pt}
\end{figure}
Table~\ref{tab:comp_freefall} shows that our model outperforms both baselines. PhysGen struggles with accurately modeling spatial relationships between objects and surfaces like ground or tables, thus often leads to physically implausible object interactions. For PISA, its best variant with depth-based reward optimizes for trajectory accuracy~(comparable L2/CD) at the cost of shape consistency~(lower IoU). In contrast, ours excels in IoU while maintaining competitive trajectory accuracy, achieving the best overall performance, which is also proved in Figure~\ref{fig:comp_freefall}.

\noindent \textbf{Ablation study.} 
We report the qualitative results from different training stages and pipelines on the synthetic subset of PisaBench in~Table~\ref{tab:abla_freefall},
where block 1 denotes the I2V base model; block 2 - 4 refer to our model and its variants in Stage I - III of training pipeline.
``Seen'' corresponds to a split of videos with objects and backgrounds seen during training, and ``Unseen'' with novel objects and backgrounds.
\textit{\textbf{1) Ablation for different training stages.}}
row 3, 5, and 8 indicate that our SFT endows the model with preliminary ability to predict objects' motion of ``free-fall'', the subsequent DPO for DiT model further steers the generated videos' distribution towards physically plausible paths, and the optimization of \encoder in the last stage improves its capability in guiding model towards higher level of physics-awareness. The qualitative results in Figure~\ref{fig:stage_real} consistently prove our pipeline's efficacy.
\textit{\textbf{2) Ablation for \encoder.}}
The comparative pipeline in row 2 and 4 is not equipped with \encoder, thus SFT and DPO are both implemented on the DiT model. Although the performance after SFT on both DiT and \encoder~(row 3) is even worse than SFT on DiT alone~(row 2), showing that single SFT cannot help \encoder learn appropriate physical representations for guiding the DiT model towards physics-awareness, DPO unlocks \encoder’s potential to extract physical information and guide the model to generate videos with better physical performance~(row 4, 5,8).
\textit{\textbf{3) Ablation for DPO strategies.}}
All strategies of DPO succeed in further improving physical accuracy on average than previous stages.
Only optimizing \encoder~(row 6) encounters difficulties in performance improvements. The model itself has not been aligned to adherence to physics before providing feedback to \encoder, preventing DPO from functioning effectively. 
Although Stage II of our training pipeline underperforms joint DPO of DiT and \encoder~(row 7), our Stage III surpasses all other methods. Joint optimization of DiT and \encoder in Stage III~(row 9) achieves comparable overall performances with our Stage III but performs worse on "unseen" split, probably because this variant with trainable DiT is more likely to overfit training data, harming the model's generalizability to novel scenarios. 

\vspace{-2pt}
\noindent \textbf{PCA analysis.} 
We also visualize the principal component analysis~(PCA) on the physical features from \encoder in Stage I and Stage III in Figure~\ref{fig:rep}. In
our Stage III physical feature maps, similarities are shown clearly for objects under the same external forces,~(green for objects in mid-air and only affected by gravity, red for objects on the ground subjected to the support force); differences are also shown more obviously between materials~(e.g., deformable object in white box has clearly distinct colors), which proves two aspects of physical understanding of our \encoder.

\vspace{-6pt}
\subsection{Generalization on General Open-world Scenarios}
\label{sec:generalization}
\vspace{-4pt}


\begin{figure*}[t]
\centering 
\includegraphics[width=1\linewidth]{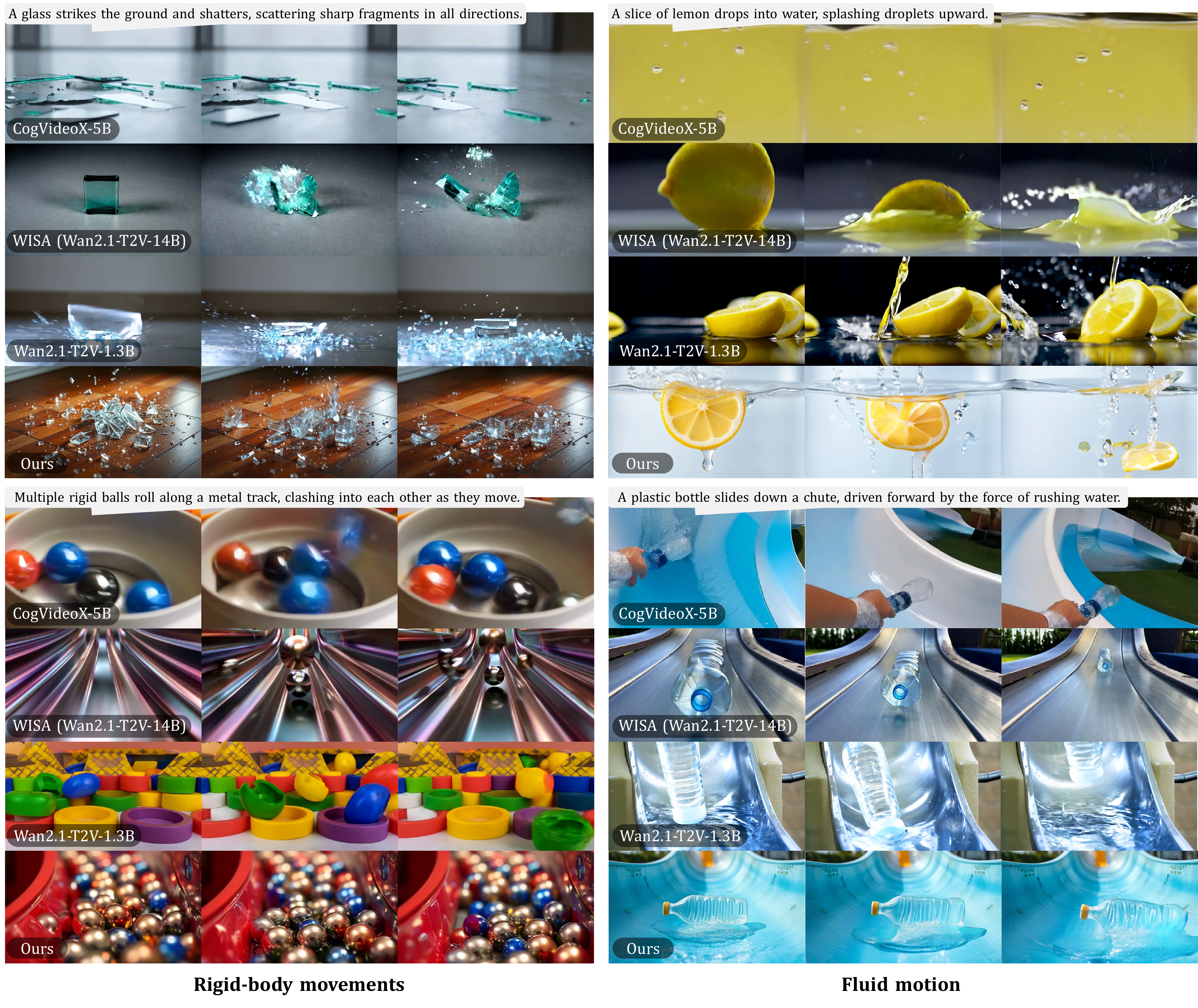} 
\vspace{-12pt}
\caption{%
    \textbf{Qualitative comparison with existing T2V models on general open-world scenarios} including objects of various materials and in different environments, validates the generalizability of our method.}
\label{fig:comp_general}
\vspace{-13pt}
\end{figure*}

\begin{figure*}[t]
\centering 
\includegraphics[width=1\linewidth]{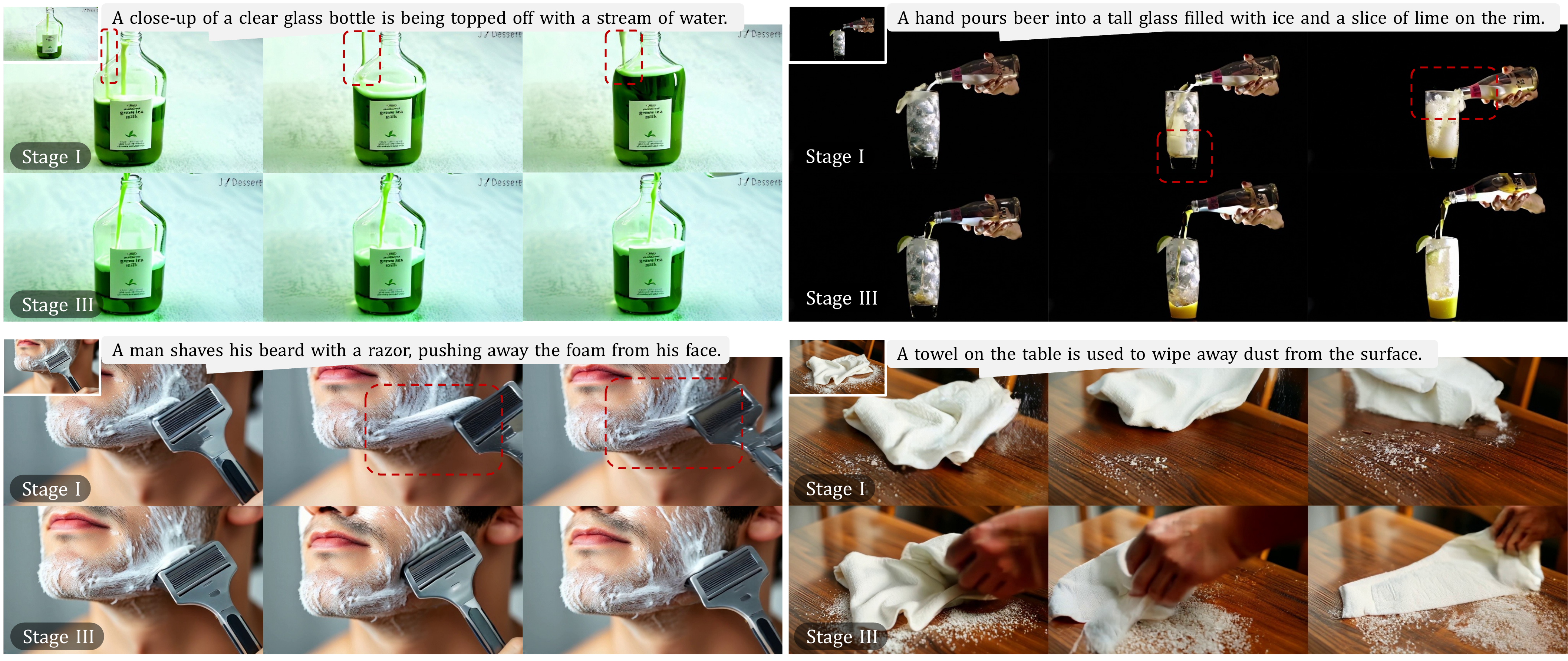} 
\vspace{-12pt}
\caption{%
    \textbf{Qualitative ablation for models in different stages} on general open-world scenarios. DPO following Stage I improves the physical coherence of model in Stage III~(e.g., fluid mechanics and gravitation).}
    
\label{fig:abla_general}
\vspace{-14pt}
\end{figure*}


\encoder demonstrates its physics-awareness for enhancing model's physical realism on the proxy task, suggesting its potential to generalize to broader open-world scenarios. 
We apply our training pipeline on a large-scale dataset~\citep{wang2025wisa} broadly covering common physical phenomena observed in real world to substantiate the generalizability of our method.

\noindent \textbf{Comparison.} We compare with two types of video generation models, general models including HunyuanVideo~\citep{kong2024hunyuanvideo}, CogVideoX-5B~\citep{yang2024cogvideox}, Cosmos-Diffusion-7B~\citep{agarwal2025cosmos}, Wan2.1-T2V-1.3B and specialized physics-focused models represented by PhyT2V~\citep{xue2024phyt2v} and WISA~\citep{wang2025wisa}. Table \ref{tab:comp_all} shows that, although our base model is surpassed by CogvideoX-5B, the base model of WISA, our final model in Stage III achieves state-of-the-art performance on both SA and PC metrics, demonstrating that our proposed method enhances the realism of generated videos physically and semantically. Our model also has a significant advantage in efficiency. It is approximately 70x faster than PhyT2V—an iterative method requiring feedback from VLM, and 8x faster than WISA. Our model generates a 5-second video in just 26 seconds on a single A800 GPU, establishing it as a highly practical solution without sacrificing physical or semantic adherence. Figure~\ref{fig:comp_general} includes qualitative comparison with existing T2V models, demonstrating our superior ability in challenging cases of both rigid-body and fluid motion.
\begin{table}[!t]
    \centering
    \begin{minipage}{0.44\textwidth}
        \centering
        \caption{\textbf{Quantitative comparison with existing video generation models} on general open-world scenarios. Our model shows superior performance in both physics-awareness and efficiency.}
        \vspace{-7pt}
        \label{tab:comp_all}
        \scriptsize
        \setlength{\tabcolsep}{3.3pt}
        \begin{tabularx}{\textwidth}{@{}l >{\centering\arraybackslash}p{2cm} >{\centering\arraybackslash}X >{\centering\arraybackslash}X@{}}

            \toprule
            Methods & Inference Time~(s) & SA~($\uparrow$) & PC~($\uparrow$) \\
            \midrule
            HunyuanVideo & 1080 & 0.46 & 0.28 \\
            Wan2.1-T2V-1.3B & 180 & 0.49 & 0.24 \\
            CogvideoX-5B & 210 & 0.60 & 0.33 \\
            Cosmos & 600 & 0.57	& 0.18 \\
            PhyT2V & 1800 & 0.61 & 0.37 \\
            WISA & 220 & \textbf{0.67} & \underline{0.38} \\
            Our base model & \textbf{23} & 0.59	& 0.29 \\
            \rowcolor{shadecolor} 
            Our final model & \underline{26} & \textbf{0.67} & \textbf{0.40} \\
            \bottomrule
        \end{tabularx}
    \end{minipage}\hfill
    \begin{minipage}{0.52\textwidth}
        \centering
        \caption{\textbf{Ablation study for \encoder} on general open-world scenarios. $v_\theta$ is DiT model, $E_{p}$ is \encoder. The results validate that DPO enables $E_{p}$ to acquire a comprehensive understanding of real-world physics and thus effectively enhances the physics awareness of $v_\theta$.}
        \label{tab:abla_all}
        \scriptsize
        \setlength{\tabcolsep}{3.3pt}
        \renewcommand{\arraystretch}{0.9}
        \begin{tabularx}{1\textwidth}{@{}lXX@{}}
            \toprule
            Methods & SA~($\uparrow$) & PC~($\uparrow$) \\
            \midrule
            Base & 0.59 & 0.29 \\
            \midrule
            SFT for $v_\theta$ & 0.63 & 0.33 \\
            SFT for $v_\theta$ + DPO for $v_\theta$ & 0.64 & 0.35 \\
            \midrule
            SFT for $v_\theta$ \& $E_p$~(Stage I) & 0.61 & 0.33 \\
            \rowcolor{shadecolor} 
            SFT for $v_\theta$ \& $E_p$ + DPO for $v_\theta$ + DPO for $E_{p}$~(Stage III) & \textbf{0.67} & \textbf{0.40} \\
            \bottomrule
        \end{tabularx}
        \end{minipage}
    \vspace{-15pt}
\end{table}

\begin{wraptable}{r}{0.4\textwidth}
\vspace{-10pt}
    \centering
    \vspace{-5pt}
    \caption{\textbf{User study for models from different stages} validates the effect of
    our training pipeline in two selected physical scenarios. Our final model shows superior ability of \encoder in Stage III in enhancing the model's physics-awareness over the base model and Stage I.}
    \resizebox{.9\linewidth}{!}
    {
    \centering
    \scriptsize
    \renewcommand{\arraystretch}{0.8}
    \setlength{\tabcolsep}{3.3pt}
    \begin{tabular}{lcc}
        \toprule
        Methods & Rigid-body movement & Fluid motion \\
        \midrule
        Base & 7.8 & 12.2 \\
        Stage I & 25.3 & 16.7 \\
        \rowcolor{shadecolor} 
        Stage III & \textbf{66.9} & \textbf{71.1}\\
        \bottomrule
    \end{tabular}
    }
    \label{tab:user_study}
\vspace{-8pt}
\end{wraptable}
\noindent \textbf{Ablation study.} We conduct ablation analysis to verify the effectiveness of our core component and strategy of training in Table~\ref{tab:abla_all}. \textit{\textbf{1) Effectiveness of \encoder:}} Compared to our base model~(row 1), our final model~(row 5) improves SA and PC scores by 0.08 and 0.11. The comparative pipeline is not equipped with \encoder, with SFT~(row 2) and the following DPO~(row 3) both implemented on the DiT model only. Such a pipeline without \encoder improves SA and PC scores by 0.05 and 0.06, proving the advantage of our proposed \encoder in successfully extracting crucial physical knowledge from the training data and using it to guide the generator toward greater physical realism, which is unattainable by simply applying SFT or DPO to the DiT model alone. 
\textit{\textbf{2) Effectiveness of DPO:}} Simply applying SFT to \encoder~(row 2 vs. row 4) does not yield an immediate benefit, suggesting that SFT alone is insufficient for \encoder to learn a useful guiding physical representation. However, DPO unlocks the potential of \encoder, allowing it to effectively translate its learned physical representations into improved generation quality in both physical commonsense and semantic adherence~(row 4 vs. row 5).
Additionally, Figure~\ref{fig:abla_general} visualizes the videos generated by our model in Stage I and III, further validating the effectiveness of DPO. \textit{\textbf{3) Effectiveness of whole training pipeline:}} Table~\ref{tab:user_study} includes human preference rates among models from different training stages in two types of real-world scenarios, showing that annotators prefer videos from Stage III model than Stage I model or I2V base model in adherence to physical laws. 

\vspace{-12pt}
\section{Conclusions}
\label{sec:conclusion}
\vspace{-8pt}
We propose \method, which learns physical representation from input image for guiding I2V model to generate physically plausible videos. We optimize physical encoder based on generative feedback from a pretrained video generation model via DPO, which proves to enhance the model's physical accuracy and demonstrate generalizability across various physical processes by injecting physical knowledge into generation,  proving its potential to act as a generic and plug-in solution for physics-aware video generation and broader applications.

\noindent \textbf{Limitations.} 
We rely on human annotators to construct preference datasets for DPO in real-world scenarios, which is costly and time-consuming. Existing AI evaluators, however, have flawed physics knowledge and inherit biases, limiting the scalability of reinforcement learning. Fortunately, our DPO training paradigm is effective even with a small amount of human-labeled data~(500 in our experiment), mitigating this limitation.




\bibliography{iclr2026_conference}
\bibliographystyle{iclr2026_conference}

\end{document}